# COSMO: Combination of Selective Memorization for Low-cost Vision-and-Language Navigation

Siqi Zhang[1]*, Yanyuan Qiao[3], Qunbo Wang[2], Zike Yan[4], Qi Wu[3], Zhihua Wei[1]†, Jing Liu[2]

[1]School of Computer Science and Technology, Tongji University
[2]Institute of Automation, Chinese Academy of Sciences
[3]Australian Institute for Machine Learning, The University of Adelaide
[4]Institute for AI Industry Research (AIR), Tsinghua University

{2211172,zhihua_wei}@tongji.edu.cn, {yanyuan.qiao,qi.wu01}@adelaide.edu.au,
qunbo.wang@ia.ac.cn, yanzike@air.tsinghua.edu.cn, jliu@nlpr.ia.ac.cn

## Abstract

*Vision-and-Language Navigation (VLN) tasks have gained prominence within artificial intelligence research due to their potential application in fields like home assistants. Many contemporary VLN approaches, while based on transformer architectures, have increasingly incorporated additional components such as external knowledge bases or map information to enhance performance. These additions, while boosting performance, also lead to larger models and increased computational costs. In this paper, to achieve both high performance and low computational costs, we propose a novel architecture with the **co**mbination of **s**elective **m**em**o**rization (COSMO). Specifically, COSMO integrates state-space modules and transformer modules, and incorporates two VLN-customized selective state space modules: the Round Selective Scan (RSS) and the Cross-modal Selective State Space Module (CS3). RSS facilitates comprehensive inter-modal interactions within a single scan, while the CS3 module adapts the selective state space module into a dual-stream architecture, thereby enhancing the acquisition of cross-modal interactions. Experimental validations on three mainstream VLN benchmarks, REVERIE, R2R, and R2R-CE, not only demonstrate competitive navigation performance of our model but also show a significant reduction in computational costs.*

## 1. Introduction

Enabling human-robot interaction through natural language instructions has been a long-standing objective in artificial intelligence, with applications ranging from home assistant robots to autonomous navigation systems in dynamic envi-ronments. In pursuit of this goal, the Vision-and-Language Navigation (VLN) task has been proposed and has emerged and garnered significant attention across multiple disciplines [2, 24, 80]. Given a natural language instruction, a VLN agent is tasked to comprehend the instruction and navigate through an unseen environment to reach the designated destination. Throughout the process, the agent accumulates a substantial amount of visual observations from the environment and should selectively retain visual contents pertinent to the given instruction.

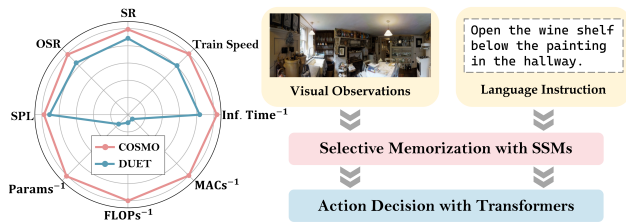

Figure 1. COSMO is a hybrid architecture that first selectively memorizes instruction-related visual observations via VLN-tailored state-space modules, then achieves precise action decision-making via Transformers. Comparisons of navigation performance and computational costs between baseline DUET [9] and COSMO on the REVERIE dataset are shown on the left, where higher values are preferred for all metrics.

Early methods [15, 66] employ recurrent neural networks to compress navigation history into one fixed-size feature vector, which would inevitably lead to information loss [14]. To tackle this challenge, Transformer-based architectures are adopted to retain historical observations into a variable-length sequence [8, 72]. For instance, the widely adopted method DUET [9] in VLN proposes a dual-scale graph Transformer architecture. It integrates a topological map to track historical and candidate nodes, which are then encoded using Transformers. Building upon DUET, many recent approaches [1, 43, 47, 48, 74] have achieved sub-stantial performance gains by incorporating external knowledge or map information. However, while these advancements enhance model performance, they also result in larger models and increased computational costs. In addition, despite the utilization of Transformers, we observe that as the length of instructions increases, the complexity of navigation path also increases, resulting in a notable decline in navigation performance of current methods.

---

*This work was done during internship at the Institute of Automation.
†Corresponding author

Recently, state space models (SSMs) [21, 22, 64] have garnered significant attention across various disciplines due to their linear computational complexity and promising potential for handling lengthy sequences. The selective SSMs [12, 20] achieve further progress by incorporating an input-dependent design that enables efficient data selection. Owing to the low computational complexity and superior long-sequence modeling capability of SSMs, developing VLN models based on SSMs appears to be a promising direction to mitigate the high computational costs and suboptimal performance on lengthy instructions that are encountered by current Transformer-based models. However, due to their inherent design for single 1-D sequence modeling, SSMs face challenges in **(i)** learning intricate spatial relationships among visual observations and performing cross-modal attention similar to Transformers. Furthermore, although SSMs outperform Transformers in many generative tasks, **(ii)** they fall short in tasks that require input selection [69]. Unfortunately, VLN is a dynamic action decision task that requires the agent to select a view from input observations as its action at each time step. Due to these two challenges, we observe that direct application of the selective SSM from Mamba [20] to VLN tasks would result in non-trivial performance degradation, as shown in Table 3.

In this context, we propose a novel architecture with the **co**mbination of **s**elective **m**em**o**rization (COSMO), a cost-efficient yet effective architecture designed for the VLN task. To address challenge **(i)**, we introduce two customized selective state space modules for VLN: Round Selective Scan (RSS) for spatial modeling, and Cross-modal Selective State Space Module (CS3) for cross-modal interaction. In contrast to recent methods that rely on multi-directional scanning for spatial-aware comprehension [49, 61, 85], RSS facilitates a comprehensive understanding of inter-token relationships within a single scan, thereby enhancing efficiency. CS3 adapts the selective mechanism to a dual-stream architecture with multimodal inputs, facilitating enhanced interaction and mutual selection between the two modalities, thereby improving its suitability for the VLN task. Regarding challenge **(ii)**, we propose a hybrid architecture that integrates the memory filtering ability of selective SSMs and the contextual selection of Transformers. RSS and CS3 are first employed for selective memorization, followed by the attention mechanism for precise action decision-making. Experiments on three popular VLN benchmarks, REVERIE [56], R2R [2], and R2R-CE [36], demonstrate competitive performance. As illustrated in Figure 1, our method not only significantly reduces the computational costs but also attains superior navigation performance compared to the DUET [9] baseline. We found that the proposed COSMO exhibits absolute improvements of $+3.83\%$ and $+2.2\%$ on SR and SPL over DUET on the REVERIE validation unseen set, $+5\%$ and $+4\%$ on SR and SPL on the R2R-CE test set. Notably, these performance gains are achieved while COSMO only obtains $15.5\%$ total parameters and incurs $9.3\%$ FLOPs compared to DUET.

In summary, the main contributions are as follows:
- We propose the combination of selective memorization (COSMO), an innovative approach that achieves low-cost VLN by combining two VLN-specified SSMs designed for selective memorization.
- We propose two customized SSMs for VLN: Round Selective Scan (RSS) to capture comprehensive inter-token relationships within a single scan, and Cross-modal Selective State Space Module (CS3) to facilitate cross-modal interactions and mutual selection.
- Extensive experiments demonstrate the effectiveness of our proposed COSMO, which achieves competitive performance while maintaining a significantly low computational cost.

## 2. Related Work

**Vision-and-Language Navigation (VLN).** VLN task is a crucial component for versatile embodied navigation agents and has gained significant research attention [4, 5, 38, 67, 80, 84]. Early approaches employ encoder-decoder frameworks [2, 6, 15, 40, 66] to remember previously visited places using recurrent states. Subsequent approaches adopt Transformer-based architecture [68] and follow the pretraining-and-finetuning strategy, resulting in substantial improvements in navigation performance [8, 27, 29, 51]. Building upon this foundation, map-based methods propose to explicitly memorize navigation history by constructing a topological map [9–11, 18, 45, 53, 71, 73], a top-down semantic map [7, 19, 31, 33], or a grid map [74]. In addition, some approaches employ the concept of the world model for future image prediction and mental planning to enhance navigation performance [35, 39, 42, 70, 75]. Furthermore, incorporating more comprehensive visual cues like depth and map information [1, 32, 47, 48] or commonsense knowledge [17, 43, 58, 60] has also been demonstrated to enhance navigation performance. Recently, NaviLLM [83] has introduced the first generalist agent capable of handling a wide range of embodied tasks by harnessing the power of large language models and leveraging vast training data. However, as performance improves, the model size expands, resulting in significantly higher computational expenses for the agent during navigation. Nev-ertheless, previous efforts have paid little attention to mitigating computation costs. VLN-PETL [59] focuses on storing minimal parameters for every downstream task and enhancing fine-tuning efficiency. In this paper, we propose COSMO, which aims to reduce computational costs while maintaining the agent's navigation performance by integrating two novel linear complexity modules.

**State Space Models (SSMs).** SSMs have demonstrated significant effectiveness in sequence modeling. HiPPO [21] captures long-term sequence dependencies by compressing inputs using high-order orthogonal polynomials. S4 [22] proposes to reduce computational and memory requirements by decomposing the structured state matrices into a low-rank and a normal term. Building upon S4, variants with different architectures are proposed [16, 23, 26, 28, 64, 65]. For instance, S5 [64] supports multi-input and multi-output and introduces efficient parallel scan algorithms. GSS [52] proposes a gate mechanism that incorporates a compacted SSM architecture. Mamba [12, 20] distinguishes itself by enhancing S4 with a selection mechanism and an input-dependent SSM layer. These studies primarily focus on demonstrating the effectiveness and efficiency of state space models in long-range and causal data modeling. Many subsequent works extend Mamba to other fields, such as vision [41, 49, 85], multi-modal [61, 81], generation [79], and robotics [54], etc. In the vision domain, the great potential of Mamba has inspired a series of works in medical image segmentation [46, 50, 62, 76, 78]. In the multimodal domain, VL-mamba [61] and Cobra [81] leverage Mamba LLM for conducting multimodal reasoning. Furthermore, several approaches are proposed for multimodal image fusion [44, 55, 77]. However, in both VL-mamba and Cobra, features from different modalities are concatenated and treated as one sequence, whereas in image fusion, two sequences of equal length are merged. Therefore, exploring the application of state space models to fuse sequences of disparate lengths across diverse modalities remains an unexplored direction. VLN tasks pose challenges due to the substantial imbalance in sequence length between visual and textual modalities [3, 8]. This discrepancy necessitates not only fine-grained modality alignment but also comprehensive intermodal interaction. In this paper, we propose two customized selective space modules specifically designed for the VLN task. The Round Selective Scan facilitates comprehensive inter-modal interactions within a single scan, and the Cross-modal Selective State Space Module adapts the selective mechanism into a dual-stream structure that enables comprehensive interaction of multimodal information in the state space.

## 3. Preliminaries

**Problem Definition.** In the standard VLN setup for discrete environments [2], the environment is an undirected naivgation graph $\mathcal{G} = \{\mathcal{V}, \mathcal{E}\}$, where $\mathcal{V} = \{V_i\}_{i=1}^K$ denotes $K$ navigable nodes, and $\mathcal{E}$ denotes connectivity edges. Given an instruction with $L$ words $\mathcal{I} = \{w_i\}_{i=1}^L$, the goal of the agent is to traverse the navigation graph according to the instruction to the goal location and find the object if required by the instruction. At each step $t$, the agent receives a panoramic view $\mathcal{O}_t$ and neighboring nodes $\mathcal{N}(V_t)$ of its current node $V_t$. $\mathcal{O}_t$ can be split into $N$ view images: $\mathcal{O}_t = \{v_i^t\}_{i=1}^N$, where $v_i$ represents the $i$-th view image of node node $V_t$. The action space $\mathcal{A}_t$ at step $t$ contains navigating to $V_{t+1} \in \mathcal{N}(V_t)$ and stopping at $V_t$.

**State Space Model.** Traditional SSMs [22, 64] can be regarded as linear time-invariant (LTI) systems that map a scalar input $x(t) \in \mathbb{R}, t = 1, ..., L$ to the output response $y(t) \in \mathbb{R}$ through a hidden state $h(t) \in \mathbb{R}^N$, where $L$ is sequence length and $N$ is state size. The continuous-time form of SSMs is often formulated as linear ordinary differential equations (ODEs):

$$h'(t) = \mathbf{A}h(t) + \mathbf{B}x(t), \quad y(t) = \mathbf{C}h'(t) + \mathbf{D}x(t) \quad (1)$$

where $\mathbf{A} \in \mathbb{R}^{N \times N}$ is the evolution matrix, $\mathbf{B} \in \mathbb{R}^{N \times 1}$ and $\mathbf{C} \in \mathbb{R}^{N \times 1}$ are the projection parameters related to system input and output, and $\mathbf{D} \in \mathbb{R}$ is the skip connection weight.

In order to be integrated into deep models, continuous-time SSMs need to be discretized. To achieve this, a timescale parameter $\mathbf{\Delta} \in \mathbb{R}$ is introduced to transform the continuous parameters $\mathbf{A}$ and $\mathbf{B}$ into discrete parameters $\bar{\mathbf{A}}$ and $\bar{\mathbf{B}}$. The zero-order hold (ZOH) method is commonly employed for the transformation:

$$\begin{aligned}\bar{\mathbf{A}} &= \exp(\mathbf{\Delta}\mathbf{A}) \\ \bar{\mathbf{B}} &= \exp(\mathbf{\Delta}\mathbf{A})^{-1}(\exp(\mathbf{\Delta}\mathbf{A}) - \mathbf{I}) \cdot \mathbf{\Delta}\mathbf{B} \approx \mathbf{\Delta}\mathbf{B}\end{aligned} \quad (2)$$

After descretization, Equation (1) can be reformulated with step size $\mathbf{\Delta}$ as:

$$h_t = \bar{\mathbf{A}}h_{t-1} + \bar{\mathbf{B}}x_t, \qquad y_t = \mathbf{C}h_t + \mathbf{D}x_t \quad (3)$$

In practice, $x_t$ is a feature vector with $D$ dimensions, and Equation (3) operates on each dimension independently. When the parameters remain constant values, Equation (1) represents a linear time-invariant (LTI) system and treats all the input tokens equally. Mamba [20] selects data by learning input-dependent parameters and surpasses the traditional SSMs:

$$\mathbf{B}_t = S_B(x_t), \quad \mathbf{C}_t = S_C(x_t), \quad \mathbf{\Delta}_t = \tau_\Delta(S_\Delta(x_t)) \quad (4)$$

where $S_B, S_C, S_\Delta$ are linear projection layers, and $\tau_\Delta$ is SoftPlus, a smooth approximation of ReLU.

## 4. Method

### 4.1. Baseline Method

We apply DUET [9] as our baseline model, which has been widely employed as a strong baseline by recent studies [18, 47, 48]. DUET adopts a dual-stream architec-ture, comprising a text encoder and a panorama encoder to extract single-model features, along with coarse-scale and fine-scale cross-modal encoders to fuse multi-modal features and learn actions at both scales. Predictions from two scales are dynamically fused as the ultimate action.

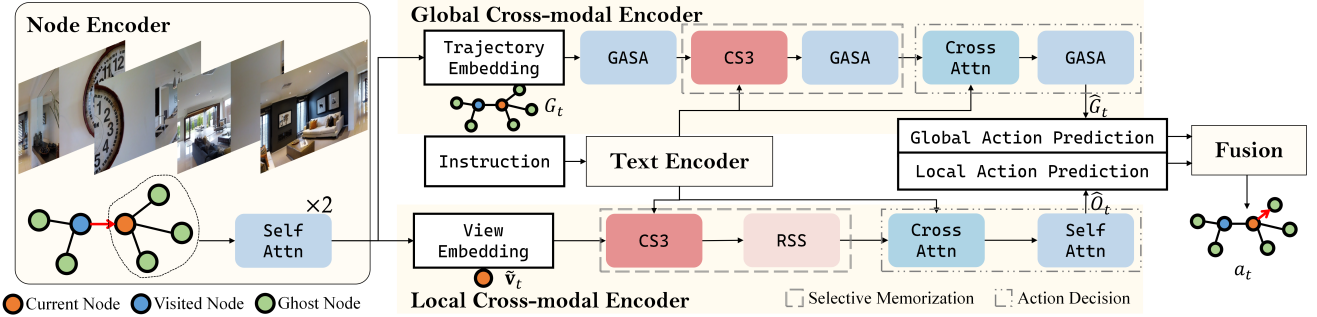

Figure 2. The overall framework of COSMO. In the Node Encoder, a topological map is constructed by compressing observations at each node to represent the node. The Global Cross-modal Encoder encodes the topological map, while the Local Cross-modal Encoder encodes the view features of the current node. RSS denotes Round Selective Scan (Section 4.2.1), CS3 denotes Cross-modal Selective State Space Module (Section 4.2.2), GASA denotes graph-aware self-attention.

**Text Encoder.** Each word in the instruction $\mathcal{I} = \{w_i\}_{i=1}^{L}$ is embedded and added with a positional embedding corresponding to the position of the word in the instruction, as well as a type embedding for the text. Subsequently, all word tokens are fed into a pre-trained language encoder to obtain word representations, denoted as $\mathcal{W} = \{\hat{w}_1, ..., \hat{w}_L\}$.

**Panorama Encoder.** At time step $t$, the agent receives panoramic observations with $N$ views $\mathcal{O}_t = \{v_i\}_{i=1}^{N}$. A pre-trained vision transformer [13] is applied to extract global representations of each view. These features are subsequently fed into a multi-layer transformer to model spatial relationships among each view.

**Cross-modal Encoders.** The coarse-scale encoder first constructs a topological map, where the mean of all views represents visited nodes and the current node, while the average of partial observations from already visited locations represents the ghost nodes. Then, for encoders at both scales, visual features (topological map at coarse-scale, current nodes at fine-scale) and textual features are fed into multiple cross-modal Transformer blocks to generate multi-modal features.

**Dynamic Action Fusion.** The output multi-modal features of two cross-modal encoders ($\hat{G}_t$ at coarse-scale, $\hat{O}_t$ as fine-scale) are first fed into separate FFNs to get action predictions. Subsequently, these predictions are dynamically fused via a learnable scalar to obtain the final action $a_t$.

### 4.2. COSMO

Given the high computational costs and suboptimal performance of recent models on lengthy instructions, it is promising to construct VLN models based on SSMs. However, since SSMs are inherently designed for 1-D sequence modeling, their ability to learn spatial and cross-modal relations is limited. Consequently, directly utilizing SSMs for VLN would result in subpar performance, as evidenced in Table 3. Therefore, we propose two selective state space modules tailored for VLN tasks: Round Selective Scan (RSS, Section 4.2.1) for spatial modeling and Cross-modal Selective State Space Module (CS3, Section 4.2.2) for cross-modal interaction. Furthermore, as illustrated in Figure 2, our COSMO adopts a hybrid architecture. Selective state space modules are first employed to facilitate selective memorization by filtering out redundant visual observations and those irrelevant to the given instructions, thereby enhancing navigation performance on lenghty instructions, as analyzed in Figure 4. Transformers are then utilized for informed action decision-making.

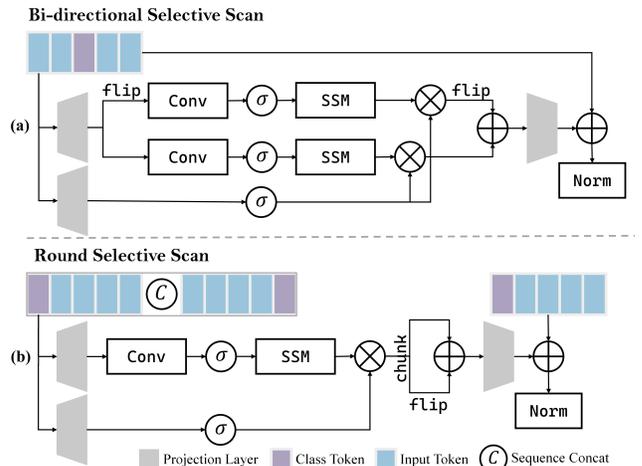

Figure 3. Structures of Bi-directional Selective Scan and our proposed Round Selective Scan (RSS).

#### 4.2.1. Round Selective Scan

Previous approaches [41, 49, 61, 85] utilize multi-directional SSMs to capture diverse token relationships, necessitating multiple rounds of scanning. For example, the bi-directional selective scan [85], as depicted in Figure 3(a), positions the class token at the center of the sequence. Then, the selective scan is performed in both forward and backward directions, enabling the class token to gather informa-tion from all other tokens within the state space and exert influence over them. This not only doubles the scanning time but also restricts the state space to include only token information preceding the current token in both scanning directions. Thus, the bi-directional selective scan still adheres to a causal mode. However, VLN agents do not require causal computation. Instead, they must comprehend the objects within each view and the spatial relationships among views in the panoramic observation.

**Algorithm 1** Cross-modal Selective State Space Module
**Input:** $x$: (B,S,D), $y$: (B,L,D)
**Output:** $x$: (B,S,D)
1: $y'$: (B,2L,E) $\leftarrow$ Linear$_y$([$y$|flip($y$)])
2: $y_{in}$: (B,2L,E) $\leftarrow$ SiLU(Conv1d($y'$))
3: $\mathbf{B}$: (B,2L,N) $\leftarrow$ Linear$_B(y_{in})$
4: $\mathbf{\Delta}$: (B,2L,E) $\leftarrow$
   $log(1 + exp($Linear$_\Delta(y_{in})$ + Parameter$_\Delta))$
5: $\mathbf{C}$: (B,1,N) $\leftarrow$ Linear$_C(x[:,0])$
6: $\bar{\mathbf{A}}$: (B,2L,E,N) $\leftarrow \mathbf{\Delta} \otimes$ Parameter$_\mathbf{A}$
7: $\bar{\mathbf{B}}$: (B,2L,E,N) $\leftarrow \mathbf{\Delta} \otimes \mathbf{B}$
8: $y_{out}$: (B,1,E) $\leftarrow$ SSM($\bar{\mathbf{A}}, \bar{\mathbf{B}}, \mathbf{C}$)($y$)[:, $-1$] $\triangleright$ Gate
9: $x_z$: (B,S,E) $\leftarrow$ SiLU(Linear$_x(x)$)
10: $x_{cross}$: (B,S,D) $\leftarrow$ Linear$_{out}(y_{out} \odot x_z)$
11: $x \leftarrow$ Norm($x_{cross} + x$)

To meet the demand of the VLN task and enhance efficiency, we propose Round Selective Scan (RSS), which captures inter-token relationships within a single scanning round. As illustrated in Figure 3(b), the input sequence $x'$ is first flipped and concatenated with the original sequence $x = [x'|$flip$(x')]$, then the scanning process is only conducted once. By having the class token positioned at both ends of $x$, all the tokens can access global information and exert influence on it. The state space already encompasses information about all the tokens when scanning the second half of $x$, facilitating more effective encoding with sufficient contextual information. After completing the selective scan, the output $y$ is divided equally in the token dimension and then reversed in the second half. The two segments are added to get the RSS result. Leveraging the hardware-aware parallel algorithm, increasing the length in the token dimension has minimal impact on both training and inference time. The RSS module not only facilitates the acquisition of comprehensive contextual information by each token but also encapsulates compressed information of the entire sequence in the class token as it corresponds to the final state in the state space.

#### 4.2.2. Cross-modal Selective State Space Module

Due to their inherent design for single-stream sequence modeling, SSMs face challenges in performing cross-modal attention similar to transformers. As a result, previous approaches in the field of multimodal learning have primarily focused on either concatenating textual and visual features into one sequence or applying them to image fusion tasks where both sequences possess an equivalent number of tokens. However, single-stream modeling is inadequate for VLN tasks [8], which is also proved in our ablation (Table 3). Therefore, we propose the Cross-modal Selective State Space Module (CS3) which adapts SSMs to a dual-stream structure better suited for VLN tasks.

As illustrated in Algorithm 1, $x$ and $y$ are the feature vectors of two modalities, S and L are their respective sequence length. Assuming the objective is to update $x$ with $y$ (and vice versa), *i.e.* extracting information from $y$ conditioned on $x$ and updating $x$ accordingly. Therefore, the state space is updated with $y$, namely constructing $\mathbf{B}$ and $\mathbf{\Delta}$ based on $y$. Same as in RSS, $y$ is first flipped and concatenated with the original sequence $y$ to enable all the tokens to access and impact the class token. The translation from the state space to the output should be guided by the target modality, namely constructing $\mathbf{C}$ based on $x$. We take the class token of $x$ to construct $\mathbf{C}$. After transfer in the state space, the output associated with the class token of $y$, which also represents the final state, functions as a gate to selectively filter relevant information from $x$. In practice, take the local cross-modal encoder in Figure 2 as an example, $x$ represents view embedding, $y$ represents instruction embedding.

#### 4.2.3. Hybrid Architecture

**Node Encoder.** The panorama encoder in DUET exclusively encodes the current node and subsequently constructs the topological map in the coarse-scale cross-modal encoder. In contrast to DUET, we first establish the topological map in the node encoder, and then encode both the current node and its unvisited neighboring nodes $\mathcal{N}(V_t) \setminus \{V_i\}_{i=1}^{t-1}$. The panorama encoder in DUET does not model the interdependencies among observations for ghost nodes, thus delegating this work to the cross-modal encoder, consequently leading to an increased requirement of parameters for the cross-modal encoder.

Formally, the topological map at time step $t$ is denoted as $\mathcal{G}_t = \{\mathcal{V}_t, \mathcal{E}_t\}, \mathcal{V}_t = \{V_i\}_{i=1}^t \cup \{\mathcal{N}(V_i)\}_{i=1}^t$ and consists of $K_t$ nodes. The topological map documents all observations of each node ($\mathbf{v_i} = \{v_j^i\}_{j=1}^{N_i}, v_j^i \in \mathbb{R}^D$ for node $V_i$ with $N_i$ observations), as well as the compressed representation of each node $\{\hat{\mathbf{v}}_\mathbf{i}\}_{i=1}^{K_t}, \hat{\mathbf{v}}_\mathbf{i} \in \mathbb{R}^D$ after undergoing encoding by the node encoder. The observations of each node are encoded using a two-layer self-attention (SA) mechanism, followed by obtaining a compressed representation through an average pooling layer:

$$\hat{\mathbf{v}}_\mathbf{i} = \text{Avg}(\text{SA}(\text{SA}(\{v_j^i\}_{j=1}^{N_i}))) \quad (5)$$

Observation of the current node is not fed into the average pooling layer. Its encoded embedding $\tilde{\mathbf{v}}_\mathbf{t} \in \mathbb{R}^{N_t \times D}$ serves as the input for the local cross-modal encoder. The compressed representations are updated to $\mathcal{G}_t$ and constructs the trajectory embedding $G_t = \{\hat{\mathbf{v}}_\mathbf{i}\}_{i=1}^{K_t}$ which is then fed into the global cross-modal encoder.

| Methods | REVERIE Dataset | | | | | | R2R Dataset | | | | | | Param (M)↓ | FLOPs (G)↓ |
|---|---|---|---|---|---|---|---|---|---|---|---|---|---|---|
| | Val Unseen | | | Test Unseen | | | Val Unseen | | | Test Unseen | | | | |
| | OSR↑ | SR↑ | SPL↑ | OSR↑ | SR↑ | SPL↑ | NE↓ | SR↑ | SPL↑ | NE↓ | SR↑ | SPL↑ | | |
| KERM [43] | 55.21 | 50.44 | 35.38 | 57.58 | 52.43 | 39.21 | 3.22 | 72 | 61 | 3.61 | 70 | 59 | 222 | 15.24 |
| BEVBert [1] | 56.40 | 51.78 | 36.37 | 57.26 | 52.81 | 36.41 | 2.81 | 75 | 64 | 3.13 | 73 | 62 | 181 | 17.71 |
| GridMM [74] | 57.48 | 51.37 | 36.47 | 59.55 | 53.13 | 36.60 | 2.83 | 75 | 64 | 3.35 | 73 | 62 | 161 | 9.53 |
| ScaleVLN [73] | 63.85 | 56.97 | 41.84 | 62.65 | 56.13 | 39.52 | 2.09 | 81 | 70 | 2.27 | 80 | 70 | 181 | 4.95 |
| RecBert [29] | 35.02 | 30.67 | 24.90 | 32.91 | 29.61 | 23.99 | 3.93 | 63 | 57 | 4.09 | 63 | 57 | 160 | 2.93 |
| AirBERT [25] | 34.51 | 27.89 | 21.88 | 34.20 | 30.28 | 23.61 | 4.01 | 62 | 56 | 4.13 | 62 | 57 | 251 | 4.14 |
| HAMT [8] | 36.84 | 32.95 | 30.20 | 33.41 | 30.40 | 26.67 | 3.65 | 66 | 61 | 3.93 | 65 | 60 | 170 | 6.29 |
| TD-STP [82] | 39.48 | 34.88 | 27.32 | 40.26 | 35.89 | 27.51 | 3.22 | 70 | **63** | 3.73 | 67 | **61** | 172 | 7.87 |
| HOP+ [57] | 40.04 | 36.07 | 31.13 | 35.81 | 33.82 | 28.24 | 3.49 | 67 | 61 | 3.71 | 66 | 60 | 160 | 2.93 |
| LANA [72] | 38.54 | 34.00 | 29.26 | 36.41 | 33.50 | 26.89 | - | 68 | 62 | - | 65 | 60 | 213 | 4.80 |
| VLN-PETL [59] | 37.03 | 31.81 | 27.67 | 36.06 | 30.83 | 26.73 | 3.53 | 65 | 60 | 4.10 | 63 | 58 | - | - |
| NaviLLM [83] | 52.27 | 42.15 | 35.68 | 51.75 | 39.80 | 32.33 | 3.51 | 67 | 59 | 3.71 | 68 | 60 | 6633 | 1011.19 |
| DUET [9] | 51.07 | 46.98 | 33.73 | 56.91 | 52.51 | 36.06 | 3.31 | 72 | 60 | 3.65 | 69 | 59 | 181 | 4.95 |
| COSMO (Ours) | **56.09** | **50.81** | **35.93** | **59.33** | **52.53** | **36.12** | **3.15** | **73** | 61 | **3.43** | **71** | 58 | **28** | **0.46** |

Table 1. Comparison with SoTA methods on both the REVERIE dataset and the R2R dataset. The grayening methods are the ones that leverage extra scene data, depth information, or external knowledge.

**Global Cross-modal Encoder.** At the global level, the agent should choose one of the ghost nodes as its next action or stop at the current node. To represent the stop action, an all-zeros vector is prepended to $G_t$. The node sequence is first passed through a graph-aware self-attention (GASA) layer [9] to enable the stop token to capture information from all other tokens. Subsequently, the encoded node and instruction features are fed into a CS3 to facilitate semantic alignment between the nodes and the instruction while effectively filtering out visual information irrelevant to the given instruction. The node sequence is fed into a GASA layer that models intra-modality interactions among nodes. Since CS3 performs semantic alignment in the state space and filtering in the feature dimension, the node and instruction features then undergo a cross-attention layer to conduct grounding at the token dimension as well as a GASA layer. The encoded node feature is denoted as $\hat{G}_t \in \mathbb{R}^{K_t \times D}$.

**Local Cross-modal Encoder.** At the local level, the agent attends to the current location $V_t$ and should select one of the views $\mathcal{O}_t$ corresponding to a neighboring node or decide to stop at the current node. Taking $\tilde{\mathbf{v}}_t$ as input, a class token is prepended to represent the stop action. The features of the current node and the instruction are fed into a cross-modal fusion module, which includes CS3 for alignment and selective memorization, RSS for broadcasting contextual visual information, cross-attention for token-level grounding, and self-attention for intra-modality interaction. The encoded view feature is denoted as $\hat{O}_t \in \mathbb{R}^{N_t \times D}$.

## 5. Experiments

### 5.1. Experimental Setup

**Datasets.** We evaluate the proposed method on R2R [2], REVERIE [56] and R2R-CE [36] datasets. R2R provides step-by-step instructions that the agent should strictly adhere to. REVERIE offers high-level instructions specifying only the destination and target object, requiring the agent to explore the environment. R2R-CE is a variant of R2R in continuous environments.

**Evaluation Metrics.** We utilize standard evaluation metrics [2], including (1) Navigation Error (NE): the average distance between the agent's final location and the destination; (2) Success Rate (SR): the ratio of successful navigation. Navigation is deemed successful if the agent stops within 3 meters of the destination; (3) Oracle SR (OSR): the ratio of tasks of which one of its trajectory nodes stops within 3 meters of the destination; (4) SR penalized by Path Length (SPL): measures both the accuracy and efficiency of navigation, which normalizes the success rate with trajectory length.

**Implementation Details.** We utilize TinyBert [34] as our text encoder, with a hidden size of 312 and an intermediate size of 1200. The state space size for RSS and CS3 are set to 16. To ensure a fair comparison, we maintain consistency with DUET [9] by employing identical input features and hyper-parameters. The optimal checkpoints are selected based on SR+SPL on the validation unseen split.

### 5.2. Comparison with State-of-the-Art

**Navigation Performance.** As shown in Table 1, we present a comparison of our COSMO with state-of-the-art (SoTA) models on REVERIE and R2R. Our model demonstrates competitive performance across the two datasets. For instance, when compared to NaviLLM on REVERIE, which incorporates a large language model and an extensive training corpus, our COSMO demonstrates superior performance with an 8.66% higher SR on the validation unseen split. Moreover, our approach exhibits better generalization capability on the test split, surpassing NaviLLM by 12.73% and 3.79% in terms of SR and SPL. In comparison to our baseline DUET, COSMO achieves a 3.83% improvement in SR on the validation unseen split and delivers competitive performance on the test unseen split.

| Methods | Val Unseen | | | Test Unseen | | |
|---|---|---|---|---|---|---|
| | OSR↑ | SR↑ | SPL↑ | OSR↑ | SR↑ | SPL↑ |
| GridMM[†] [74] | 61 | 49 | 41 | 56 | 46 | 39 |
| BEVBert[†] [1] | 67 | 59 | 50 | 67 | 59 | 50 |
| Seq2Seq* [36] | 40 | 32 | 30 | 36 | 28 | 25 |
| WPN [37] | 40 | 36 | 34 | 37 | 32 | 30 |
| CM2* [19] | 42 | 34 | 28 | 39 | 31 | 24 |
| MGMap* [7] | 48 | 39 | 34 | 45 | 35 | 28 |
| CMA[†] [30] | 52 | 41 | 36 | 49 | 38 | 33 |
| VLNBERT[†] [30] | 53 | 44 | 39 | 51 | 42 | 36 |
| DUET[†] [9, 74] | **58** | **47** | 39 | 50 | 42 | 36 |
| COSMO[†] (ours) | 56 | **47** | **40** | **55** | **47** | **40** |

Table 2. Comparison on the R2R-CE dataset. ∗ methods apply a forward-facing camera with a 90° HFOV instead of panoramic images. † methods apply the same waypoint predictor [30].

| Method | OSR↑ | SR↑ | SPL↑ | Inf. Time (s)↓ |
|---|---|---|---|---|
| Mamba | 39.19 | 32.25 | 21.50 | 8.94 |
| Bi-Mamba | 42.15 | 35.61 | 24.33 | 9.42 |

Table 3. Performance of single-stream structure models where Mamba layers are directly applied to VLN tasks.

| Components | | REVERIE Val Unseen | | | |
|---|---|---|---|---|---|
| # RSS | CS3 | OSR↑ | SR↑ | SPL↑ | Inf. Time(s)↓ |
| 1 Mamba | ✓ | 52.34 | 47.20 | 32.04 | 10.46 |
| 2 Bi-Mamba | ✓ | **56.80** | 50.75 | 34.77 | 11.38 |
| 3 ✓ | Bi-Mamba | 53.14 | 46.95 | 31.40 | 10.80 |
| 4 ✓ | ✓ | 56.09 | **50.81** | **35.93** | 10.64 |

Table 4. Ablation results of the RSS and CS3 module.

For the results on the R2R dataset, our method achieves comparable performance to DUET while exhibiting a 2% improvement in SR on the test split. Additionally, COSMO outperforms DUET in NE, achieving an error reduction of 0.16m and 0.22m on the validation unseen split and test split respectively. Moreover, when compared to KERM, which incorporates external commonsense knowledge to assist navigation, our method surpasses it by 1% SR on both splits, and achieves an error reduction of 0.07m and 0.18m on the validation unseen split and test split. The slight decrease in our SPL may be attributed to the selection of the optimal checkpoint based on SR+SPL. COSMO also surpasses VLN-PETL in all metrics, particularly excelling in SR with an enhancement of 8% on the test split.

Table 2 presents results on the R2R-CE dataset. COSMO surpasses DUET by 5% and 5% in terms of SR and SPL on the test split. Additionally, it surpasses GridMM on the test split, showcasing superior generalization capability.

**Computational Costs.** In addition to the competitive navigation performance, it is worth emphasizing that COSMO stands out as a method of remarkably low computational cost. As showcased in Table 1, we calculate the number of parameters (Param) and computation cost (FLOPs) of all methods. FLOPs is calculated by the fvcore library. Considering that the input data has an impact on the results, a batch of synthetic data is constructed for fair comparison: the batch size is 1, the instruction consists of 40 tokens, the current node has 3 neighboring nodes, and the agent has visited 6 nodes. It is worth noting that the grayening methods introduce additional inputs upon DUET. While these methods enhance performance, they also impose a heavier computational burden. This paper aims to optimize the model structure to minimize computational costs while maintaining competitive navigation performance. As shown by the results, COSMO exhibits remarkably low computational complexity, only requires 9.3% of FLOPs compared to DUET. Moreover, our model is characterized by a modest parameter count of 28M, representing merely 15% of the parameters employed in DUET.

### 5.3. Ablation Study

Extensive experiments are conducted to evaluate the key designs of COSMO. Results are reported on the validation unseen split of the REVERIE dataset. Besides navigation performance, we also report inference time. To mitigate the impact of navigation steps on inference time, we take the ground-truth path as navigation history and set batch size to 32. Then evaluate the inference time required for one-step navigation on the validation unseen split of REVERIE.

**Indispensability of Dual-stream Structure.** We demonstrate that the vanilla Mamba is inappropriate for VLN tasks, directly applying Mamba to VLN would result in significant performance degradation. The results are shown in Table 3. Specifically, we replace the CS3 and cross-attention layers with Mamba layers, where textual and visual tokens are concatenated into a single sequence. The inclusion of textual information in the state space is crucial when encoding visual tokens, thus the textual sequence is concatenated in front of the visual sequence. The GASA and self-attention layers are also substituted by the Mamba layers. To ensure fairness, the state space size is set to 16, consistent with our RSS and CS3 implementations. Moreover, considering the remarkable performance achieved by bi-directional selective scan [85], referred to as Bi-Mamba hereafter, we also evaluate its effectiveness by substituting the vanilla Mamba layers with Bi-Mamba layers. It can be seen that the direct application of Mamba to the VLN task without any modifications yields a final success rate of 32.35%, which is 14.73% lower than the previous SoTA method DUET and 18.56% lower than our proposed

| # | Architecture | OSR↑ | SR↑ | SPL↑ |
|---|---|---|---|---|
| 1 | DUET-mini | 51.75 | 46.61 | 30.49 |
| 2 | `Trans + SSM` | 54.52 | 46.58 | 31.38 |
| 3 | `Trans + Trans` | **56.69** | 49.47 | 31.10 |
| 4 | `SSM + SSM` | 49.28 | 41.92 | 27.61 |
| 5 | `SSM + Trans` | 56.09 | **50.81** | **35.83** |

Table 5. Ablation on the designation of model architecture.

COSMO. Even with the improved Bi-Mamba, the success rate is only 3.26% higher compared to that of Mamba.

**Superiority of RSS.** To verify the validity of RSS module, we replace it with the vanilla Mamba layer and Bi-Mamba layer. As shown in Table 4, after replacing RSS with Mamba layer, SR experiences a decrease of 3.61%, while SPL decreased by 3.89% (#1 vs. #4). Upon replacement with a Bi-Mamba layer, it is still 1.16% lower than RSS in SPL (#2 vs. #4) and requires longer inference time.

**Superiority of CS3.** We replace the CS3 module with a Bi-Mamba layer where the textual and visual tokens are concatenated into one sequence. The results are shown in Table 4, the removal of the CS3 module impacts the navigation performance to a large extent, resulting in a decrease of 3.86% in SR and a decline of 4.53% in SPL (#3 vs. #4). These results suggest that the CS3 module outperforms the Bi-Mamba layer in learning cross-modal interactions.

**Necessity of Hybrid architecture.** To substantiate the efficacy of employing SSMs prior to Transformers, we conduct structural ablation as presented in Table 5. The Selective Memorization in Figure 2 is referred to as `SSM`, and the Action Decision is denoted as `Trans`. Results in row #3 and #4 demonstrate the indispensability of a hybrid architecture. Comparison between row #2 and #5 proves the rationality of conducting `SSM` before `Trans`. Given that the only structural difference between row #3 and DUET lies in the design of the node encoder, we reduce the scale of DUET in row #1 to match the layer and dimension configurations of COSMO, resulting in DUET-mini. It is also initialized with TinyBERT for fair comparison. The comparison between row #1 and row #3 validates the rationality of our node encoder.

### 5.4. Quantitative and Qualitative Analysis

**Quantitative Analysis.** To validate the efficacy of COSMO in selecting relevant memories, we compare the performance across instructions of varying lengths as illustrated in Figure 4. Specifically, we categorize the instructions in R2R and REVERIE validation unseen split based on their lengths and compare the navigation success rates of our COSMO, the baseline model DUET, and SoTA model BEVBert on these instructions. As instruction length increases, navigation tasks become progressively more challenging, resulting in a general decline in performance across three mod-els. Notably, COSMO demonstrates a more substantial improvement over DUET on longer instructions. For instance, COSMO achieves a 7.42% higher SR compared to DUET when instructions in REVERIE exceed 30 words. Additionally, although BEVBert generally outperforms COSMO, their performances are comparable on R2R long instruction (> 40 words). COSMO even surpasses BEVBert by 6.01% on REVERIE long instructions (> 30 words).

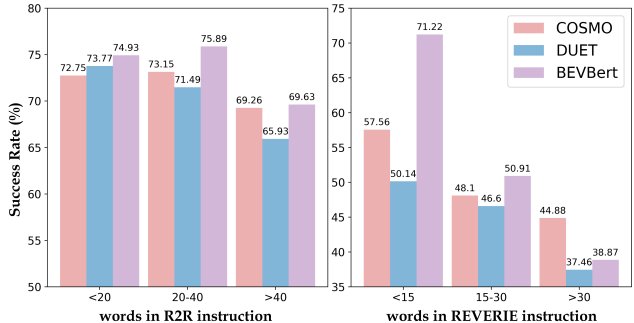

Figure 4. Comparison of navigation success rate under instructions of varying lengths in R2R (left figure) and REVERIE (right) validation unseen split.

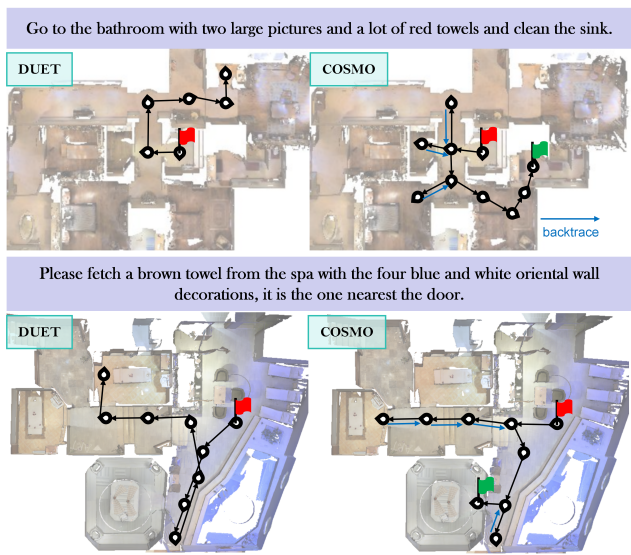

Figure 5. Predicted paths of DUET and COSMO. The agent starts from the red flag and stops at the green flag.

**Qualitative Analysis.** We visualize the predicted paths of DUET and COSMO as illustrated in Figure 5. In the upper example, DUET is incorrectly oriented and fails to backtrack, whereas COSMO explores all potential rooms that may contain a bathroom through backtracking. In the bottom example, both models pass by the target room as it is concealed. DUET ends up in a spa room without blue and white decorations, whereas COSMO exits the incorrect spa room before locating the target room.

# 6. Conclusion

In this paper, we introduce COSMO that achieves low-cost VLN with the combination of two novel SSMs designed for selective memorization. We developed two VLN-tailored SSM modules, Round Selective Scan (RSS) and Cross-modal Selective State Space Module (CS3), to optimize the navigation process in dynamic multimodal contexts. The experiments conducted on REVERIE, R2R and R2R-CE demonstrate superior performance over existing methods, particularly in handling long instructions. In addition to the performance gains, COSMO also significantly reduces computational costs.

# COSMO: Combination of Selective Memorization for Low-cost Vision-and-Language Navigation

## Supplementary Material

## 7. Experimental Setups

### 7.1. Datasets

All the datasets are built upon the Matterport3D [5] environment which contains 90 photo-realistic houses. Each house is annotated with a navigation graph, wherein the agent's movement is exclusively confined to traversing along the interconnected edges of nodes.

**R2R** [2] provides step-by-step instructions. The houses are divided into 4 sets: 61 houses for training, which are annotated with 14039 instructions; 11 and 18 houses for validation and testing in unseen environments, respectively, which are annotated with 2,349 and 41,73 instructions. Among the 61 houses for training, 56 houses are also utilized for validation in seen environments. The instructions provided in the R2R dataset describe each action the agent should take during navigation, such as "Walk straight toward the bar with the chairs. Turn left and go straight until you get to three tables with chairs. Turn left and wait near the couch". The R2R dataset consists of 21,567 navigation instructions across 7,189 paths and 10,800 panoramic views within 90 sizable real-world indoor settings. Instructions in this dataset average 29 words in length.

**R2R-CE** [36] transfers 77% of R2R paths into continuous environments, resulting a total of 5,611 paths and an average path length of 9.89m. Each instruction contains an average of 32 words. Agents have a chassis radius of 0.1m and are allowed to slide along obstacles.

**REVERIE** (Remote Embodied Referring Expression) [56] provides coarse-grained instructions. The instructions are usually concise and mainly describe the destinations and target objects, such as "Go to the bathroom with a bronze bathtub and bring me the towel above the bathtub". The training set contains 60 houses and 10,466 instructions, the validation seen set contains 46 houses and 1,423 instructions, the validation unseen set contains 10 houses and 3,521 instructions, the test unseen set contains 16 houses and 6,292 instructions. It comprises 21,702 instructions, with each instruction averaging 18 words in length. Although the trajectories in REVERIE align with that in R2R, the task presents a significantly higher level of difficulty due to the absence of explicit action guidance in the instructions, necessitating active exploration but the agent to locate the destination. Given that coarse-grained instructions bear closer resemblance to real-world scenarios, recent research has predominantly focused on the dataset.

|  | DUET | KERM | COSMO |
|---|---|---|---|
| OSR↑ | 51.07 | 55.21 | **56.09** |
| SR↑ | 46.98 | 50.44 | **50.81** |
| SPL↑ | 33.73 | 35.38 | **35.93** |
| Params(M)↓ | 181 | 222 | **28** |
| FLOPs(G)↓ | 4.95 | 15.24 | **0.46** |
| MACs(G)↓ | 4.74 | 15.04 | **0.34** |
| Inf. Time(s)↓ | 13.20 | 526.33 | **10.64** |
| Train Speed (sample/s)↑ | 29 | 2 | 36 |

Table 6. Comparison of navigation performance and computational costs between DUET, KERM, and COSMO on the validation unseen set of the REVERIE dataset.

### 7.2. Training Details

**R2R.** Following previous works [8, 9, 29], we employ augmented data [27] for pre-training. The model undergoes 100k steps of pre-training with a batch size of 64, followed by fine-tuning for 20k steps with a batch size of 8.

**R2R-CE.** We transfer the model pretrained on the R2R dataset to continuous environments through the Habitat Simulator [63]. The model is finetuned with a batch size of 16 and a learning rate of 1e-5 for 30 epochs.

**REVERIE.** Following DUET [9], we incorporate augmented data generated by the speaker model during the pre-training phase. The model is pre-trained for 100k steps with a batch size of 32, then finetuned for 20k steps with a batch size of 8.

## 8. More Comparisons

Tab. 6 presents the values in the radar chart on the left side of Fig. 1 in the paper, along with the comparison with KERM. Considering that all the metrics pertaining to computational cost aim for lower values indicating better performance, reciprocal transformations of these metrics are taken in Fig. 1 in the paper. `FLOPs` is calculated by the fvcore[1] library. `MACs` is calculated by the thop[2] library. As illustrated in Section 5.3, inference time denotes the time required for one-step navigation on the REVERIE validation unseen split. We also report training speed (referred to as Train Speed in the table). It denotes the number of samples trained per second by the model with a batch size of 32 on a single A6000 GPU. It prefers higher value.

---
[1] https://github.com/facebookresearch/fvcore.git
[2] https://github.com/Lyken17/pytorch-OpCounter.git

| Components | | REVERIE Val Unseen | | |
|---|---|---|---|---|
| # | RSS | CS3 | OSR↑ SR↑ SPL↑ | |
| 1 | Self-Attn | ✓ | 53.71 48.82 32.38 | |
| 2 | ✓ | Cross-Attn | 50.21 44.42 30.96 | |
| 3 | ✓ | ✓ | **56.09 50.81 35.93** | |

Table 7. Ablation results of the RSS and CS3 module.

| Size | OSR↑ | SR↑ | SPL↑ |
|---|---|---|---|
| 8 | 56.01 | 50.24 | 34.36 |
| 16 | **56.09** | **50.81** | **35.83** |
| 32 | 54.27 | 48.54 | 33.57 |

Table 8. Ablation on the state space size of RSS and CS3.

## 9. More Ablations

**Superiority of RSS and CS3.** Table 3 in the paper presents the ablation results of the RSS and CS3 modules without altering the hybrid architecture. To further demonstrate the efficacy of these two modules, additional ablation results are provided in Tab. 7. RSS is replaced with self-attention in row #1, resulting in a decrease of 2.0% in SR and 3.6% in SPL. This indicates that RSS effectively captures contextual relationships among tokens while efficiently compressing information into the class token. CS3 is replaced with cross-attention in row #2, leading to a significant decrease of 6.39% in SR and 4.97% in SPL. This not only highlights the necessity of employing a hybrid architecture but also demonstrates the proficiency of CS3 in modeling the interaction between modalities and their mutual selection.

**State space size.** Table 8 compares different state space sizes in our RSS and CS3. It can be observed that inadequate state space leads to insufficient retention of navigation history, while an excessively large state space results in the inclusion of redundant or noisy information during selection process.

## 10. Discussions

**Correlation between Instruction Length and Complexity of Navigation Tasks.** As the length of instructions increases, navigation tasks tend to become increasingly complex. We quantify the complexity of navigation through the length of ground-truth path. The ground-truth paths in both R2R and REVERIE exhibit a length distribution ranging from 4 to 7. Table 9 presents the average ground-truth path lengths associated with instructions of varying lengths. Table 10 presents the average instruction length associated with varying lengths of ground-truth path.

**Complementary to Existing Methods.** Given our focus on enhancing the fundamental model structure, we employ DUET [9] as the baseline model, which is widely recognized as a strong benchmark in recent literature. Our proposed RSS and CS3 can be integrated with other SoTA methods such as incorporating external knowledge in KERM [43], advancing local perception in BEVBert [1] and BSG [47]. These improvements are orthogonal to our structural enhancement. For example, we integrate the BEV features from BEVBert into COSMO, resulting in SR=71, SPL=62 on the R2R test split with only 16% parameters.

| | R2R Val Unseen | | | RVR Val Unseen | | |
|---|---|---|---|---|---|---|
| Instr len | < 20 | 20 − 40 | > 40 | < 15 | 15 − 30 | > 30 |
| GT path len | 5.73 | 6.03 | 6.24 | 5.85 | 5.99 | 6.12 |

Table 9. Average ground-truth path length associated with instructions across different length intervals.

| | R2R Val Unseen | | | | RVR Val Unseen | | | |
|---|---|---|---|---|---|---|---|---|
| GT path len | 4 | 5 | 6 | 7 | 4 | 5 | 6 | 7 |
| Instr len | 16.7 | 23.9 | 25.8 | 29.2 | 16.9 | 17.6 | 18.7 | 19.4 |

Table 10. Average instruction length associated with varying ground-truth lengths.

**Advantages of CS3 over Mamba.** As illustrated in Table 4, the performance enhancement of CS3 over Bi-Mamba is substantial ($+3.86\%$ in SR and $+4.53\%$ in SPL). This is attributed to the effective modal alignment facilitated by the dual-stream architecture of CS3. Although Mamba has been effectively utilized in the multimodal domain, such as VL-Mamba [61] and Cobra [81], these models perform modal alignment prior to inputting visual and textual features into Mamba language models. In contrast, VLN models, like HAMT [8] and DUET [9] along with their variants, employ cross-modal attention mechanisms for both multi-modal alignment and interaction. This distinction is a key reason why SSMs cannot be directly applied to VLN tasks. As illustrated in Equ(3) and Equ(4), the influence of the input token $x_t$ at time $t$ on the state space is controlled by matrix $\mathbf{B}_t$, while the resolution of the input is determined by $\mathbf{\Delta}_t$. When tokens in the input sequence originate from different modalities and are not aligned, it is evidently inappropriate to apply the same strategy ($S_B$ and $S_\Delta$) for controlling their impact on the state space and the sampling frequency. In this context, we propose CS3 as a dual-stream selective SSM. As illustrated in Algorithm 1, for instance, $x$ represents visual features, and $y$ represents textual features. Now the objective is to utilize the textual features to update the visual features. Thus, the input to the state space is $y$, while the output of state space acts upon $x$. That is to say, input matrix $\mathbf{B}$ and resolution matrix $\mathbf{\Delta}$ should be derived from $y$, whereas the output matrix $\mathbf{C}$ should be derived from $x$. This design facilitates effective alignment and interaction between multi-modal features.

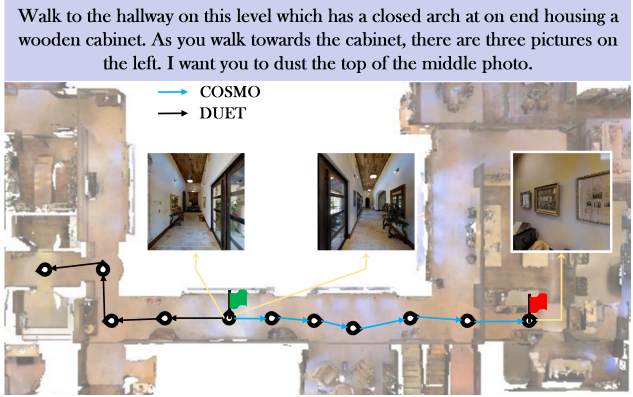

Figure 6. Predicted paths of DUET and COSMO on REVERIE validation unseen set. The red flag denotes the correct endpoint.

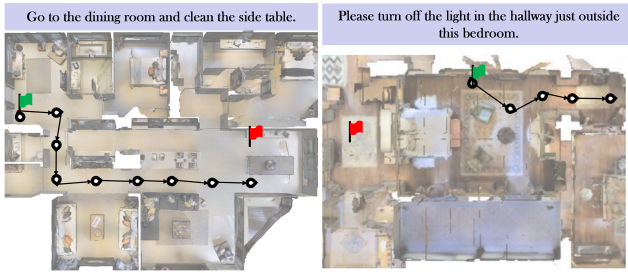

Figure 7. Failure cases of COSMO on REVERIE validation unseen set.

## 11. More Qualitative Examples

We visualize the predicted paths of DUET and COSMO in Figure 6. The instruction requires navigating to the end of a hallway featuring a closed arch. Given that the starting point is midway along a lengthy corridor, it is crucial to accurately identify the closed arch. DUET failed to select the correct direction, which ultimately led to its inability to locate the target destination. In contrast, COSMO successfully identified the direction containing the closed arch, demonstrating its ability to accurately ground the objects as described in the instruction within the environment. Consequently, COSMO find the three paintings mentioned.

We visualize two failure cases in Figure 7. In the left example, COSMO successfully located the dining room. However, it was unable to navigate closer to the side table. In the right example, COSMO was able to identify the light in the hallway, but the presence of two corridors surrounding the room led to ambiguity in the instructions, resulting in an error.